\documentclass[conference]{IEEEtran}
\IEEEoverridecommandlockouts
\usepackage{cite}
\usepackage{amsmath,amssymb,amsfonts}
\usepackage{algorithmic}
\usepackage{graphicx}
\usepackage{textcomp}
\usepackage{float}
\def\BibTeX{{\rm B\kern-.05em{\sc i\kern-.025em b}\kern-.08em
T\kern-.1667em\lower.7ex\hbox{E}\kern-.125emX}}
\begin{document}

\title{Instance Map based Image Synthesis with a Denoising Generative Adversarial Network}
\author{\IEEEauthorblockN{Ziqiang Zheng, Chao Wang, Zhibin Yu{*}, Haiyong Zheng, Bing Zheng}
\IEEEauthorblockA{College of Information Science and Engineering}
{Ocean University of China}\\
*Corresponding author: Zhibin Yu
Email: yuzhibin@ouc.edu.cn}

\maketitle

\begin{abstract}
Semantic layouts based Image synthesizing, which has benefited from the success of Generative Adversarial Network (GAN), has drawn much attention in these days. How to enhance the synthesis image equality while keeping the stochasticity of the GAN is still a challenge. We propose a novel denoising framework to handle this problem. The overlapped objects generation is another challenging task when synthesizing images from a semantic layout to a realistic RGB photo. To overcome this deficiency, we include a one-hot semantic label map to force the generator paying more attention on the overlapped objects generation. Furthermore, we improve the loss function of the discriminator by considering perturb loss and cascade layer loss to guide the generation process. We applied our methods on the Cityscapes, Facades and NYU datasets and demonstrate the image generation ability of our model.
\end{abstract}

\begin{IEEEkeywords}
Generative Adversarial network (GAN), Denoising, Image-to-image translation
\end{IEEEkeywords}

\section{Introduction}

\begin{figure}[t]
\centering
\includegraphics[width=\linewidth]{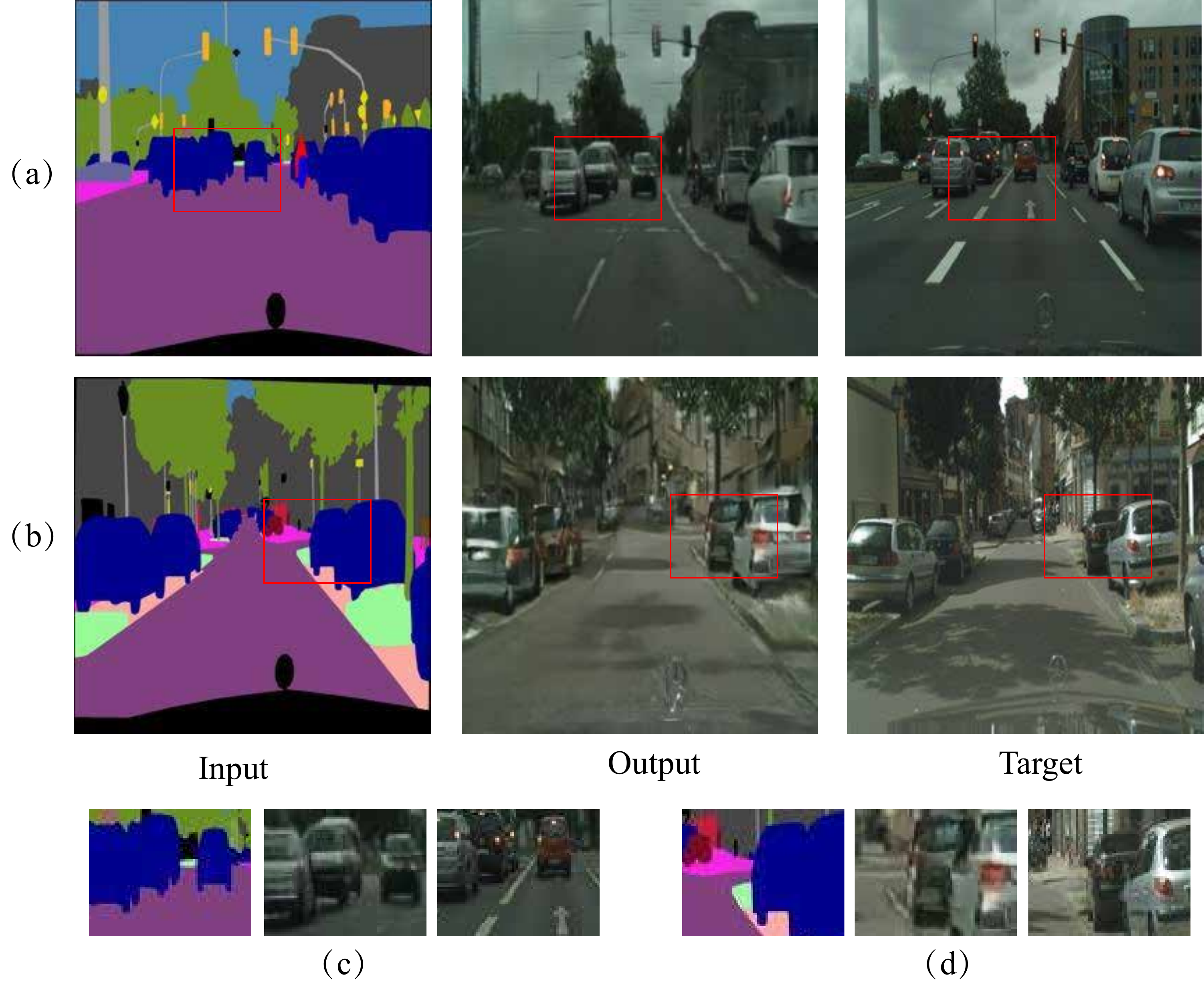}
\caption{Some image-to-image translating examples from semantic layouts to photorealistic images. (c) and (d) are local enlarged images from (a) and (b), respectively.}
\label{fig:demo}
\end{figure}

Image-to-image translation has been made much progress due to the successful of image generation and synthesizing. As a sub research topic of image translation, image-to-image translation task need the model not only to understand the image contents but also to convert the source images to the target images under some certain rules. Many image processing problems such as semantic segmentation, color restoration, etc. can be considered as image-to-image work. Since Generative Adversarial Network (GAN) was presented by Goodfellow in 2014\cite{goodfellow2014generative}, GAN is proved to be very efficient in image generation and synthesizing tasks in many research fields\cite{zhu2017unpaired}\cite{zhao2016energy}\cite{mirza2014conditional}\cite{press2017language}. Soon researchers show that the extension of GAN can not only generate images from random noise \cite{nguyen2016plug}, but also achieve the image-to-image translation between different domains \cite{isola2016image}. It's highly desirable if we could build a model which is able to create realistic images based on some simple graphics (e.g. semantic layouts). Actually, lots of works about GAN prove that GAN is a good candidate to handle the image-to-image translation work. But GAN based models are still will weak on precise generation tasks. How to organize the translated image contents precisely and enhance the generated image equality is still a challenging work. Take Fig.\ref{fig:demo} as an example, it would be a challenging task if we want to create realistic overlapped cars only based on an adhesive semantic layout. In order to achieve this goal, the image-to-image translation model should be smart enough to understand the relationship and the position of each car in the space domain.

In this paper, we proposed a novel idea for generating realistic images from the semantic layout precisely. We take advantages from pix2pix, Cascade Refinement Network (CRN) and Inception-ResNet to design our model \cite{isola2016image} \cite{chen2017photographic}\cite{szegedy2017inception}. The cascade layer loss from CRN can efficiently control the contents organizing while the adversarial structure can generate realistic sharp samples. It's believed that the residual blocks can be used to build very deep neural networks \cite{He2015}. It can also help alleviate the gradient vanish and gradient explode. We also consider the skip connection for the encoder-decoder network, which is derived from the U-net\cite{ronneberger2015u}. Specially, we add skip connections between the convolution layers before residual blocks and deconvolution layers after the residual blocks. 

In order to generate the complex objects precisely, we compute the instance map of complex objects (such as the cars in the Cityscapes\cite{Cordts2016Cityscapes} and the furniture in the NYU \cite{silberman2012indoor} datasets) from the semantic layouts. And we concatenate the instance car map with the raw semantic layout to provide additional information to our model. Inspired by InfoGAN\cite{chen2016infogan}, we develop a denoising framework by adding some auxiliary random noise to the input images. Our model will try to remove the noise during the generation process and synthesis a clear image. This operation can enhance the model stochasticity and increase the robustness in image generation stage. 
\begin{figure*}[t]
\centering
\includegraphics[width=\linewidth]{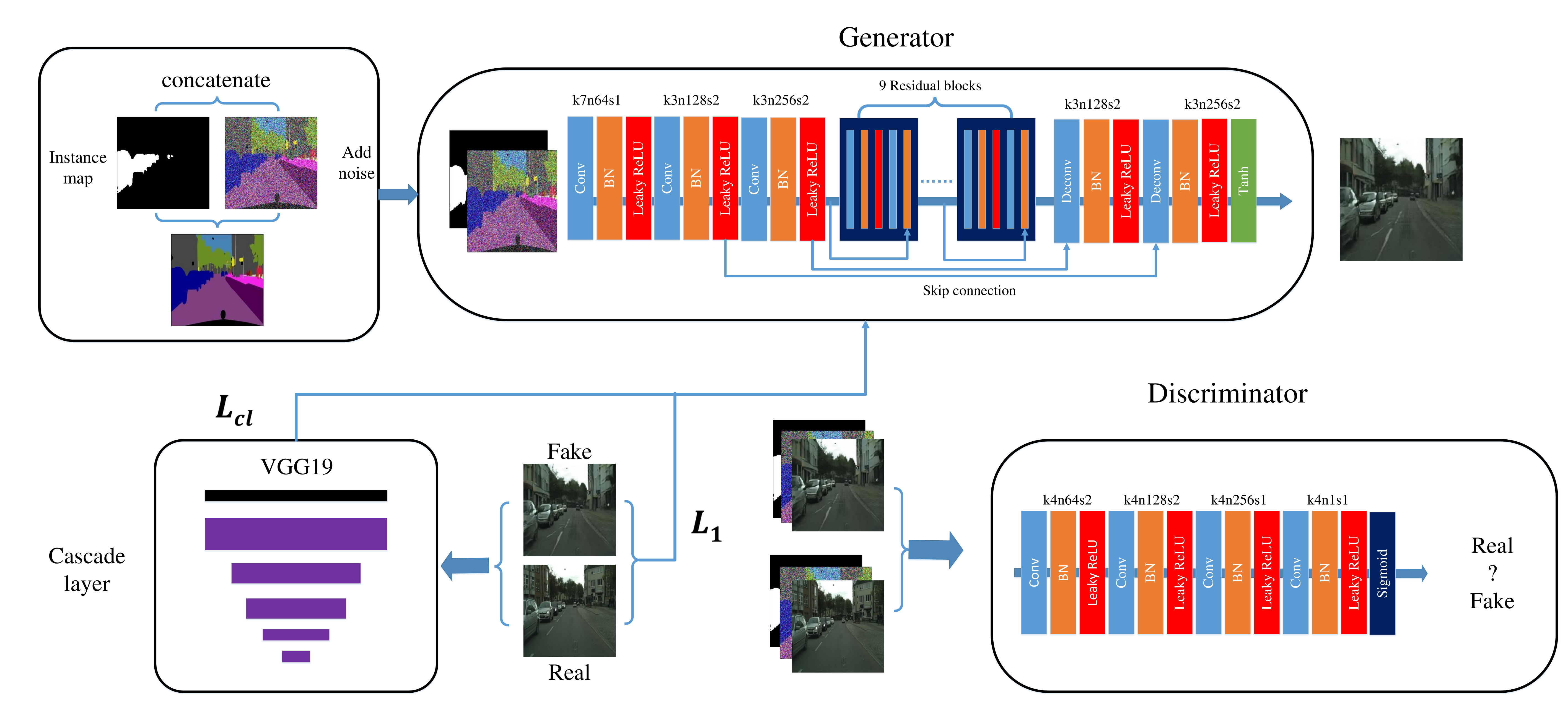}
\caption{The framework of our model.}
\label{fig:architecture}
\end{figure*}
To further improve the generation, we follow the perturbed loss defined by \cite{denton2016semi}. The perturbed loss can improve the ability of the discriminator and push the generator to create more realistic images. The detail of our framework is displayed in Fig. ~\ref{fig:architecture}.
In this paper we make the following contributions:
\begin{itemize}
\item We develop a new framework to including residual blocks, cascade layer loss and perturbed loss to synthesis images from a semantic layout to a realistic RGB image. 
\item We develop a denoising process enhance the model stochasticity and increase the robustness on image-to-image translation stage.
\item In order to generate complex objects precisely, we include an instance map to force the generator focus on the confusion parts and improve the final generation performance.
\end{itemize}
\section{Related work}
\subsection{Residual networks}
Residual networks, which are developed by \cite{He2015}, is proved to be efficient on image classification tasks. Some computer vision tasks such as image translation and image editing also use residual blocks as a strong feature representation architecture\cite{he2016identity}\cite{he2016deep}. Residual networks are neural networks in which each layer consists of a residual block $f_{i}$ and a skip connection bypassing $f_{i}$. Layers in residual networks include multiple convolution layers. The residual blocks build connections between different layers
\begin{equation}
y_{i}=f_{i}(y_{i-1})+y_{i-1} 
\end{equation}
Let $y_{i-1}$ and $y_{i}$ denote input and output of the $i$th residual block separately. Where $f_{i}$ is some sequence of convolutions, batch normalization\cite{ioffe2015batch}, and Rectified Linear Units (ReLU) as nonlinearities. The shortcut connections can transmit gradients and propagated errors in very deep neural networks.

\subsection{Image-to-image translation}
There exists a large body of work on supervised representation learning. Early methods were based on the stacked auto-encoders or restricted Boltzman machines\cite{hinton2006fast}. Researchers use pairwise supervision to achieve the image feature representation using the auto-encoders\cite{zhu2014multi}\cite{ kan2014stacked
}. And the Image-to-image translation problems are often regarded as feature representation and feature matching work. 
The researchers have already taken significant steps in the computer vision tasks, while deep convolution neural networks(CNNs) becoming the common workhorse behind a wide variety of image processing problems. For the style translation\cite{gatys2015neural}, Gatys et al applied a VGG-16 network to extract features. Other methods based on CNNs use deep neural networks to capture feature representations of images and applied them to other images\cite{johnson2016perceptual
}\cite{simonyan2014very}. They try to capture the feature spaces of scattering networks and proposed the use of features extracted from a pre-trained VGG network instead of low-level pixel-wise error measures\cite{bruna2015super}.
\subsection{GANs}

Generative adversarial networks (GANs)\cite{goodfellow2014generative} aim to map the real image distribution by forcing the synthesized samples to be indistinguishable from natural images. The discriminator try to differentiate synthesized image from real images while the generator aims to generate plausible images as possible to fool the discriminator. The adversarial loss has been a popular choice for many image-to-image tasks\cite{zhang2017age}\cite{wang2016generative}\cite{sangkloy2016scribbler}\cite{pathak2016context}.

Recent research related with GANs are mostly based on the work of DCGAN (deep convolutional generative adversarial network)\cite{radford2015unsupervised}. DCGAN has been proved to learn good feature representation from image pixels in many research. And the deep architecture has shown fantastic effectiveness of synthesizing plausible images in the adversarial networks. And conditional generative adversarial networks (cGAN) which developed by \cite{mirza2014conditional} offers a solution to generate images with auxiliary information. Previous works have conditioned GANs on discrete labels, text\cite{reed2016generative} and images. The application of conditional GANs on images processing contains the inpainting \cite{yeh2017semantic}, image manipulation guided by user constrains \cite{zhu2016generative}, future frame prediction \cite{mathieu2015deep}, and so on. Researchers applied the adversarial training to the image-to-image translation, in which we use the images from one domain as inputs to get translated images of another domain \cite{karacan2016learning}\cite{kaneko2017generative}\cite{dong2017semantic}\cite{isola2016image}. The recent famous method for image-to-image translation about GAN is pix2pix\cite{isola2016image}. It consists of a generator $G$ and a discriminator $D$. For image translation task, the objective of the generator $G$ is to translate input images to plausible images in another domain, while the discriminator $D$ aims to distinguish real images from the translated images. The supervised framework can build the mapping function between different domains. Towards stable training of GAN, Wasserstein generative adversarial network (WGAN)\cite{arjovsky2017wasserstein} replace Jensen-Shannon divergence by Wasserstein distance as optimization metric, recently some other researchers improved the training using gradient penalty method.\cite{gulrajani2017improved}.

\section{Proposed methods}
In this paper, we use the residual blocks and convolution neural networks as the generators. Inspired by the U-net, we use the skip connections between feature maps in the generator to build the deep generator network. The residual block can help alleviate the gradient vanishing and gradient explosion problems. 
In order to generate more photorealistic images and balance the training, we have added some noise to the input images and found the noise can help improve the ability of generator.
\subsection{Architecture of our models}
As described in Fig. \ref{fig:architecture}, we use three convolution-Batchnorm-ReLu\cite{ioffe2015batch} to get smaller size feature map, then we use nine residual blocks to capture feature representations. Following these residual blocks, we use two deconvolution layers to get high resolution images. We use the Batchnorm layers and Leaky ReLU after both the convolution operation and deconvolution layers. The final outputs have been normalized from -1 to 1 using the Tanh activation to generate a $256 \times 256$ image outputs. We build the skip connections between the convolution layers and deconvolution layers to allow low-level information transmit. Such a network requires that all information flow pass through all the layers, including the bottleneck. 

All the filters in convolutions have kernel size 4 and stride 2. The residual blocks are using kernel size 3 and stride 1. To learn the difference between fake and real components, we add the perturbed loss as a regularization scheme. The Fig. ~\ref{fig:architecture} shows the architecture of our model. Let $k4n64s2$ denotes the convolution layer have 64 convolution filters with kernel size 4 and stride 2. 

We use the Adam optimizer of learning rate 0.0002 for both discriminator and generator. All the experiments have taken 200 epochs. And following the pix2pix method, we have applied random jittering by resizing the $256 \times 256$ input images to $286 \times 286$ and randomly cropped back to size $256 \times 256$. And the number of our model parameter is about one fourth as pix2pix and one eighth as Cascade Refinement Network (CRN) . We use less parameter to get better results.

\subsection{Perturbed loss}
The original GAN loss is expressed as
\begin{equation}
\begin{split}
\mathcal{L}(G,D)&=\mathbb{E}_{x,y\sim P_{data}(x,y)}\left[\log D(x,y)\right]+ \\
&\!\!\!\!\!\!\!\!\!\!\!\!
\mathbb{E}_{x\sim P_{data}(x)}\left[\log (1-D(x,G(x,z)))\right].
\end{split}
\label{eq:cgan}
\end{equation}

Let $x$ denotes the semantic input and $z$ denotes the noise. As a result of concatenating the input image and generated images in the Fig. ~\ref{fig:architecture}, $D(x,G(x,z)))$ shows the predict output of the discriminator. Ideally $D(x,y)=1$ and $D(x,G(x,z)))=0$ when G(x) denotes the generated samples. To make the discriminator more competitive, we use the perturbed training to improve the generations. Following Gulrajani and Denton et al.'s work \cite{denton2016semi} \cite{gulrajani2017improved}, we define the perturbed loss $\mathcal{L}_{p}$ as:
\begin{align}
\mathcal{L}_{p} &= {E}_{x\sim P_{data}(x)}\left[\log (1-D(x,G(\hat{x})))\right]\\
\hat{x} &= \alpha G(x,z) +(1-\alpha) y
\end{align}
where $\hat{x}$ represents the mixture of the synthesized fake image $G(x,z)$ and the target image $y$. Only a portion of $\hat{x}$ is from target image. And $\alpha$ is a random number from 0 to 1 following an uniform distribution. Here we can regard $\hat{x}$ as perturbed sample. The object of an perturbed sample $\hat{x}$ indicates which we changes the output of the generated sample in a desired way. We add this perturbed loss to improve the training of GANs and the quality of synthesized samples. We compute the perturbed loss $\mathcal{L}_{p}$. The $\mathcal{L}_{p}$ computes the distance between fake images and target images and encourage the discriminator to distinguish different components of the mixed images. It can help improve the ability of discriminator to distinguish the fake images.

\subsection{Noisy adding and one-hot instance map}
Gaussian noise $z$ are usually used in the traditional GANs as an input to the generator \cite{goodfellow2014generative} \cite{radford2015unsupervised}. As mentioned in Isola et al.'s work \cite{isola2016image}, how to enhance the stochasticity of their pix2pix method is an important question. We increase the stochasticity of our model by adding the Gaussian noise to the input images. And we found the additional noise can actually help the generator synthesize images stably. And we have taken experiments using various standard deviations to try to found out the influence the noise. 

\begin{figure}[t]
\centering
\includegraphics[width=\linewidth]{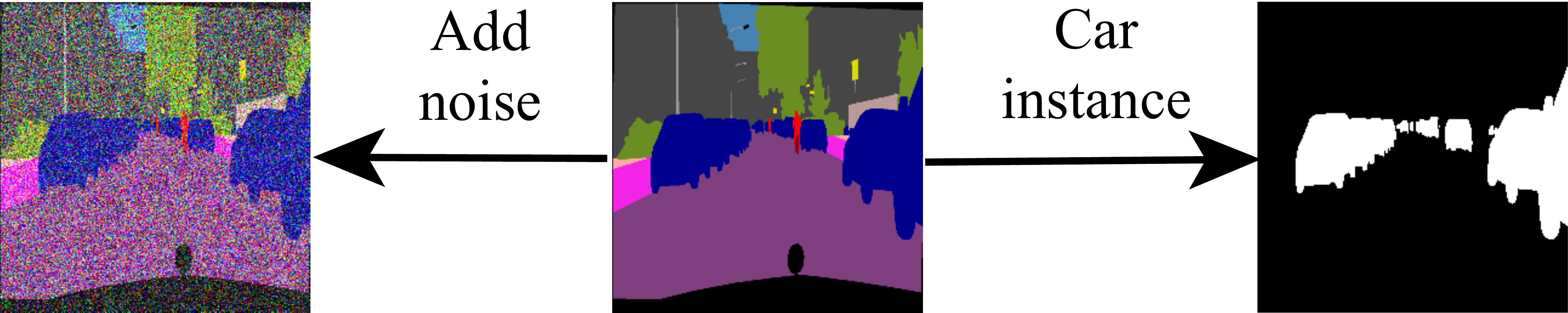}
\caption{The procedure of adding noise and creating an instance map.}
\label{fig:water}
\end{figure}

As mentioned in Fig. \ref{fig:demo}, it's a challenging work when we want to create a realistic image only based on a semantic label map while the cars are overlapped. We compute the instance map of cars from the semantic input and concatenate them as the final input. We hope this operation can force our generator to focus on the boundary and structure information of overlapped objects. First we extract the car segmentation scene and make the background black. And the Fig. ~\ref{fig:water} shows how we got the instance map from the semantic labeled images and how we added the noise. Please note that we also concatenate the instance map and generated images for the discriminator. 

\subsection{Cascade layer loss}
In order to make the synthesized images more clear and perform better, we consider Chen et al.'s work and include an extra pre-trained VGG-19 network called CRN to provide the cascade loss in our hybrid discriminator\cite{chen2017photographic}. Fig. \ref{fig:architecture} shows the detail of cascade loss. The feature maps of convolutional networks can express some important information of images. The CRN can provide extra criteria for our model to distinguish real images from fake samples. The cascade loss is considered as a measurement of similarity between the target and the output images. We initialize the network with the weights and bias of the pre-trained VGG network on the Imagenet dataset\cite{krizhevsky2012imagenet}. Each layer of the network will provide cascade loss between the real target images and the synthesized images. We use the first five convolutional layers as a component of our discriminator. 
\begin{equation}
\label{eq5}
\mathcal{L}_{c}(\theta)=\sum^{N}_{n}{\lambda}_{n}||{\Phi}_{n}(x)-{\Phi}_{n}(G((x,z);\theta))||_{1}
\end{equation}
Following the definition mentioned in Eq. \label{eq5}, here $y$ denotes the target image and $(G(x,z))$ is the image produced by the generator $G$; $\theta$ is the parameters of the generator $G$; $ {\Phi}_{n}$ is the cascade response in the $n_{th}$ level in the CRN. We choose the first 5 convolutional layers in VGG-19 to calculate the cascade loss. So we have $N=5$. Please note the loss ${L}_{cascade}(\theta)$ mentioned in Eq. \ref{eq5} is only used to train the parameter $\theta$ of the generator G. The CRN is a pre-trained network. The weights of CRN will not be changed during we train the generator $G$. The parameter ${\lambda}_{n}$ controls the influence of cascade loss in the $n_{th}$ layer of CRN.

The cascade loss can provide the ability to measure the similarity between the output and the target images under $N$ different scales and enhance the ability of discriminator. It's believed that the cascade loss from the higher layers control the global structure; the loss from the lower layers control the local detail during generation. Thus, the generator should provide better synthesized images to cheat the hybrid discriminator and finally improve the synthesized image quality.


\subsection{Objective}
The goal is to learn a generator distribution over data $y$ that matches the real data distribution $P_{data}$ by transforming an observed input semantic image $x\sim P_{data}(x)$ and a random Gaussian noise $z\sim P(z)$ into a sample $G(x,z)$. This generator is trained by playing against an adversarial discriminator $D$ that aims to distinguish between generated samples from the true data distribution (real image pair $\{x,y\}$) and the generator's distribution (synthesized image pair $\{x,G(x,z)\}$). As mentioned in Isola et al.'s work, location information is really important while synthesizing images. So we still use the L1 loss to the objective in our network to force the generator to synthesize images with considering the details of small parts. 
\begin{equation}
\mathcal{L}_{1}=\left\| y -G(x,z) \right\|_{1}.
\end{equation}
The L1 loss is the fundamental loss of our generator which guarantee the image-to-image generation abilities.

Our final objective is
\begin{equation}
G^{*} = \arg\underset{G}{\min}\;\underset{D}{\max}\mathcal{L}(G,D)+ \gamma \mathcal{L}_{l1} + \theta \mathcal{L}_{p} + \sigma \mathcal{L}_{cl}
\end{equation}
where $ \mathcal{L}_{l1}$ means the real-fake L1 loss, $\gamma$, $\sigma$ and $\theta$ are hyper parameters to balance our training. We use greedy search to optimize the hyper parameters with $\gamma=100, \theta=1, \sigma=1$ in our experiments.

\section{Experiments}
To validate our methods, we conduct extensive quantitative and qualitative evaluations on Cityscapes, Facades \cite{tylevcek2013spatial} and NYU datasets \cite{silberman2012indoor}, which contain enough object variations and are widely used for image-to-image translation analysis. We choose the contemporaneous approach (pix2pix) as the baseline \cite{isola2016image}, and we also compare our model with CRN \cite{chen2017photographic}, which is also a state of art method on image-to-image generation task. Results by the two compared methods are generated using the code and models released by their authors. Following their method, we use 400 images of Facades dataset for training and 100 for testing.

On the image-to-image tasks, we use some image quality scores (R-MSE, SSIM, MSE and P-NSR) to evaluate the generations of used methods. For these supervised translation method, output images with lower R-MSE and MSE are better while images with higher P-NSR and SSIM higher are better.

\subsection{NYU dataset and Facades dataset}
\begin{table}
\begin{center}
\caption{The image equality evaluation results of different methods on the Nyu2 dataset.}
\label{table:nyu2}
\begin{tabular}{|c|c|c|c|c|}
\hline
Method & P-SNR &MSE &R-MSE &SSIM\\
\hline\hline
Ours & 10.8562 & 0.08753& 0.2911 & 6.4610\\
CRN & 8.4967 & 0.15183 &0.3827 & 7.5927\\
Pix2pix & 10.8545 & 0.08801 &0.2916 & 6.3742\\
\hline
\end{tabular}
\end{center}

\end{table}

\begin{figure*}[t]
\centering
\includegraphics[width=\linewidth]{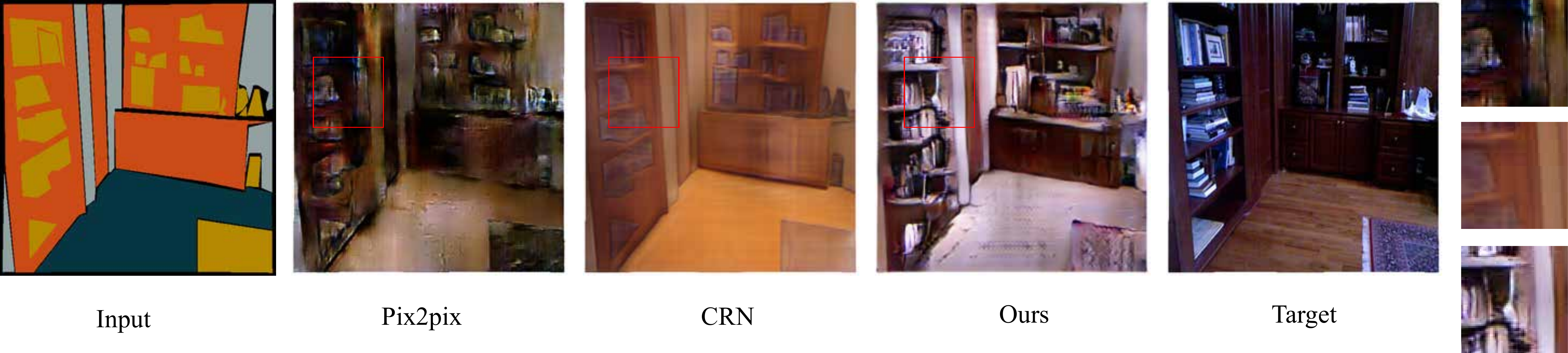}
\caption{Generation results with an indoor image from the NYU dataset\cite{silberman2012indoor}. }
\label{fig:nyu2-compare}
\end{figure*}

\begin{figure}[t]
\centering
\includegraphics[width=\linewidth]{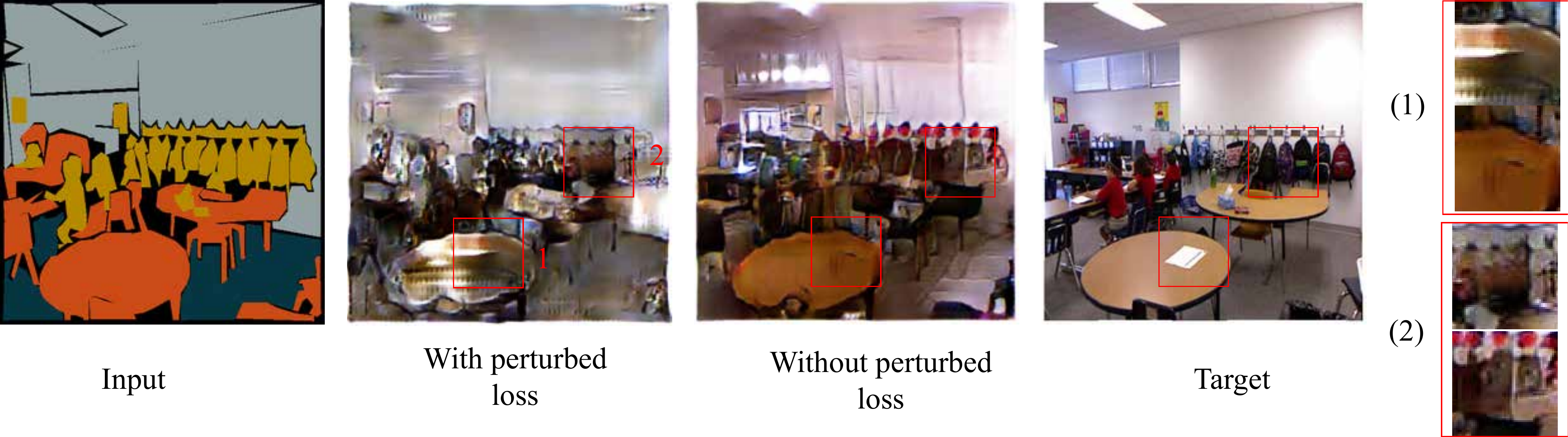}
\caption{The evaluation results of the perturbed loss on the NYU dataset. }
\label{fig:lp-nyu}
\end{figure}

In order to test the generative ability of our model, we use the RGB images from NYU dataset \cite{silberman2012indoor} and get the semantic layouts from Ivaneck's work \cite{ivaneckydepth}. One challenge on this dataset is that sematic labels are quite limited. In this dataset, the semantic layouts only contain five different classes (furniture, structure, prop, floor and boardlines). Note different type of objects can be labeled as the same class (e.g. Beds and bookshelves are both labelled as ''furniture''). Through this way, we got 795 five-category semantic images from the training dataset of raw Nyu dataset\cite{silberman2012indoor}, and 700 for training, 95 for testing. In this case, it is difficult for the generator to organize the contents and generate plausible results. We regard it as a challenging task for this one-to-many translation. So we compute the instance map of these objects and concatenate the instance map with the semantic layouts, which can provide our generator more information and force the model to pay attention to the overlapped objects. The Fig. \ref{fig:nyu2-compare} shows one generation example on NYU dataset. We can find that our result is much better than the other two methods. The relationship between books and shelves are well modeled. On the other hand, we evaluated the effectiveness of the perturbed loss $\mathcal{L}_{p}$. The Fig. ~\ref{fig:lp-nyu} shows the comparison results, which illustrate that the perturbed loss can help generate better results. 

The Fig. ~\ref{fig:nyu2-compare} shows the comparing results using the three methods on NYU dataset. The picture generated by pix2pix performs well on generating some simple contents such as floor and the wooden furniture. But pix2pix fail to generate books on the bookshelf. The CRN method fail to generate images with enough details. Although the image-to-image tasks can be considered as a one to many problem, we still measure four kinds of image quality scores on these three methods for reference. The Table.~\ref{table:nyu2} shows the evaluation results of the three methods. The CRN is better at generating images without structure errors. That's why CRN has higher SSIM score. Comparing with the pix2pix method, we have higher SSIM and lower MSE error. Our methods have better results than the other two methods.

\begin{figure}[t]
\centering
\includegraphics[width=\linewidth]{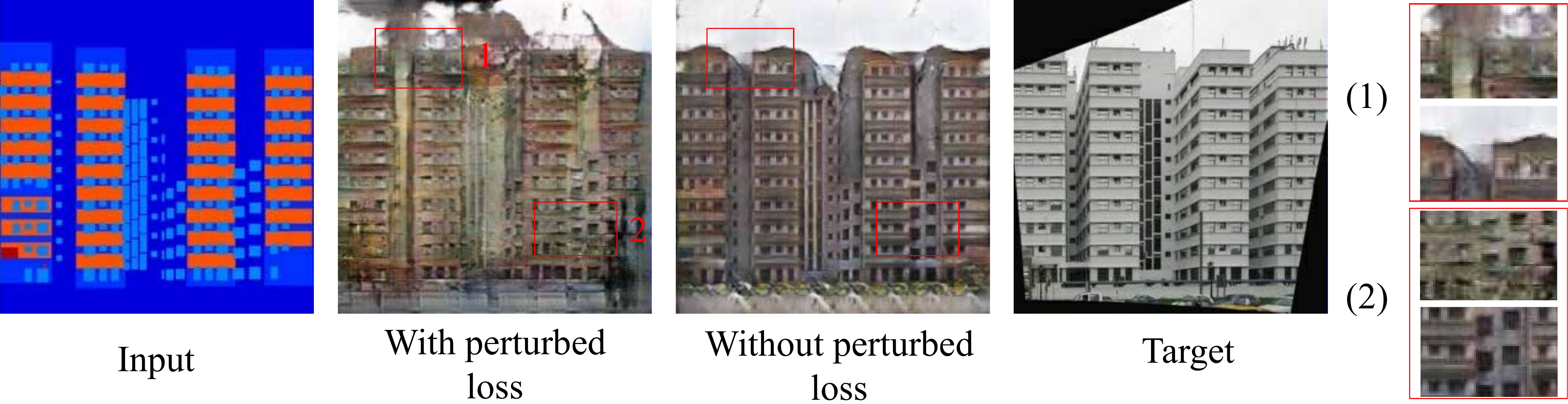}
\caption{The evaluation results of the perturbed loss on the Facades dataset. }
\label{fig:lp-facades}
\end{figure}

\begin{figure}[t]
\centering
\includegraphics[width=\linewidth]{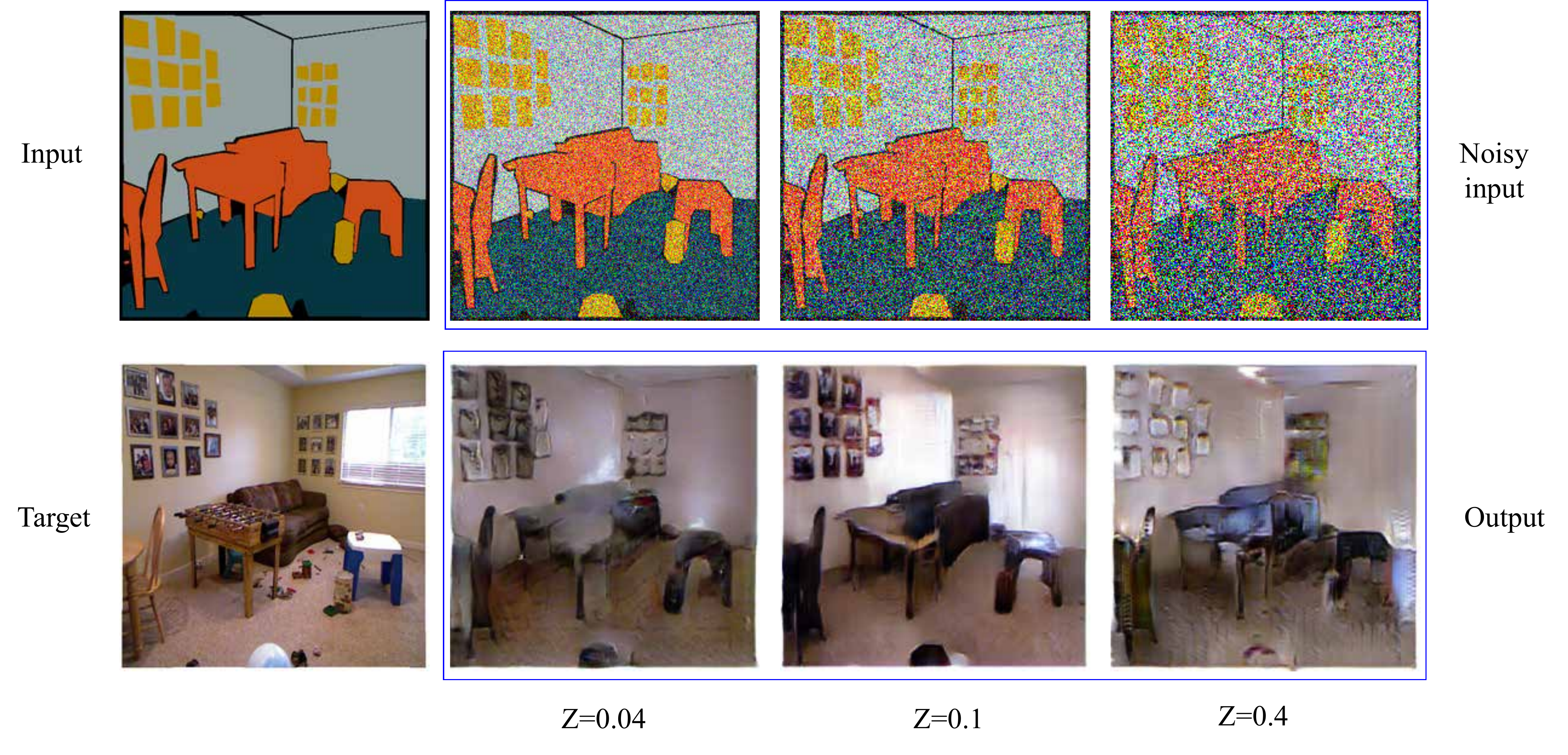}
\caption{The evaluation results of noisy inputs on the NYU dataset. }
\label{fig:noise-nyu}
\end{figure}

We evaluate the effectiveness of the perturbed loss $\mathcal{L}_{p}$ on both NYU and Facades datasets. As the Fig. ~\ref{fig:lp-facades} shows, the perturbed loss can actually reduce the blur parts and help the generator create more plausible images with more detail information. On the other hand, we evaluate the performance of denoising process and compare our methods with the pix2pix method in the Fig. ~\ref{fig:facades-noise}. The result of pix2pix has some meaningless color blocks as shown in the red boxes. The results in Fig. ~\ref{fig:facades-noise} show that the proper noise can help alleviate such phenomenon. We also evaluate the effectiveness of noise in the NYU dataset in the Fig. ~\ref{fig:noise-nyu} and get similar performance of Fig. ~\ref{fig:facades-noise}. Proper noisy inputs make the generation process more robust.

\begin{figure}[t]
\centering
\includegraphics[width=\linewidth]{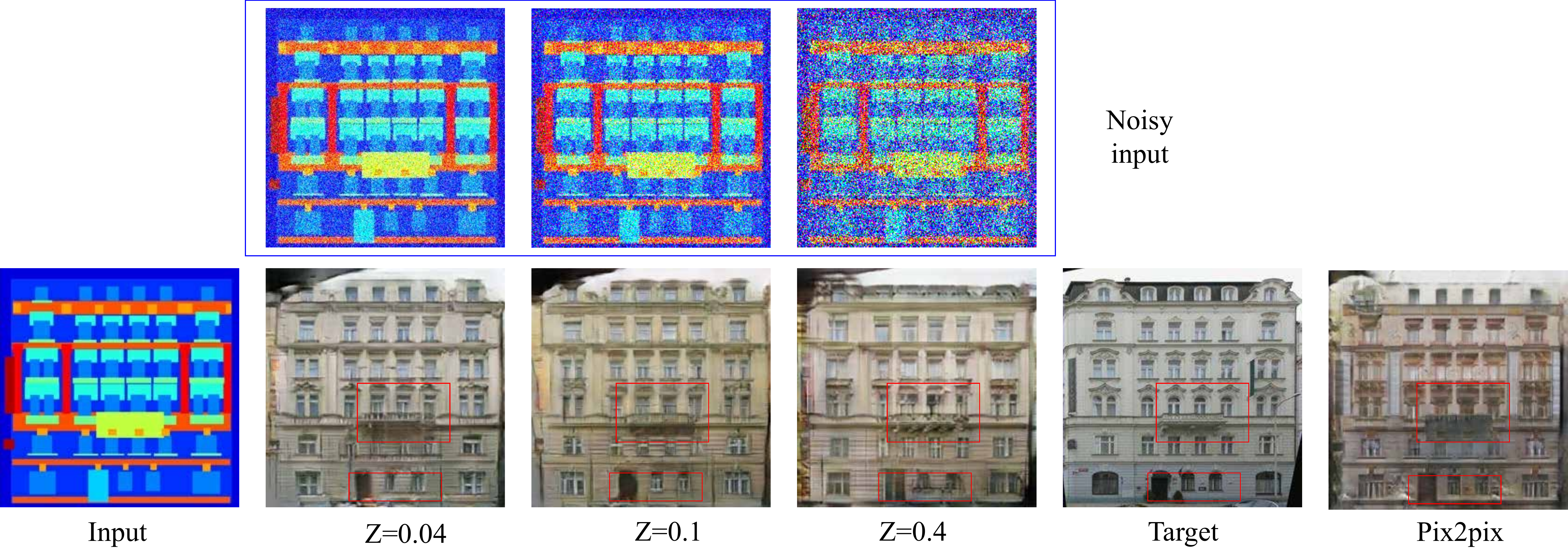}
\caption{Different scale of noise induces different quality of results on Facades dataset. We get best results with deviation 0.1}
\label{fig:facades-noise}
\end{figure}

\begin{figure}[t]
\centering
\includegraphics[width=\linewidth]{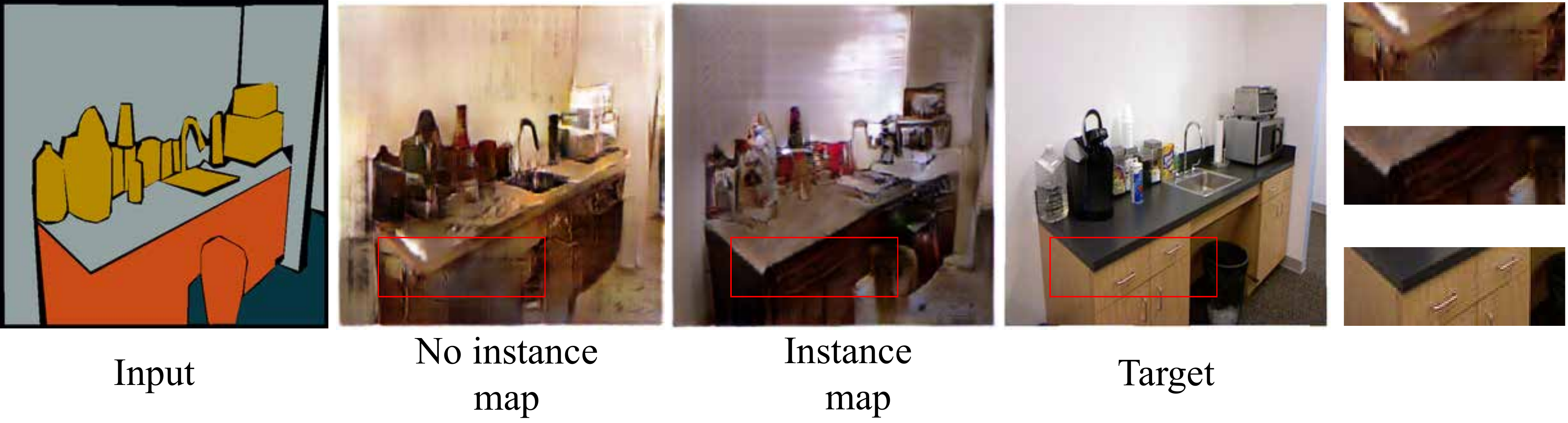}
\caption{The evaluation results of instance map on the NYU dataset. }
\label{fig:facades-noise}
\end{figure}

\begin{figure}[t]
\centering
\includegraphics[width=\linewidth]{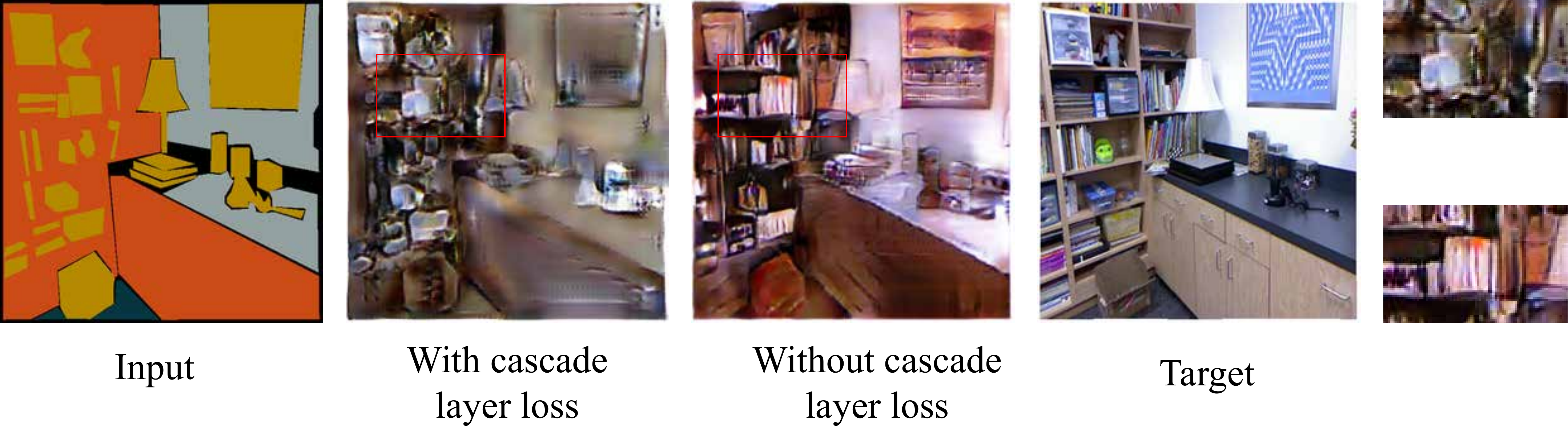}
\caption{The evaluation of the cascade layer loss on the NYU dataset.}
\label{fig:layer-nyu}
\end{figure}
\begin{figure}[t]
\centering
\includegraphics[width=\linewidth]{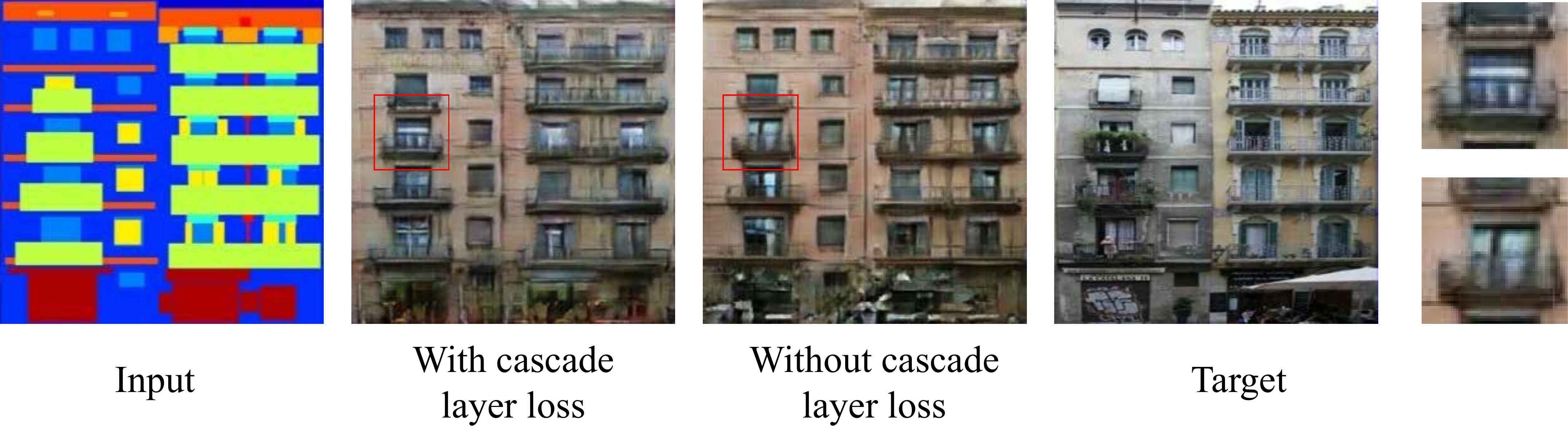}
\caption{The evaluation of the cascade layer loss on the Facades dataset.}
\label{fig:layer-facades}
\end{figure}

We also evaluate the cascade layer loss for our model. The Fig. ~\ref{fig:layer-nyu} shows the comparing results on NYU dataset. The generations without cascade layer loss are blurry and the color blocks are dirty and wrong. And we also evaluate this loss in the Facades dataset, we can see the results in the Fig. ~\ref{fig:layer-facades}. And we didn't compute the instance map for the Facades dataset as the result of that this dataset is easy and doesn't have the complex objects.

\subsection{Cityscapes dataset}

\begin{figure*}[]
\centering
\includegraphics[width=\linewidth]{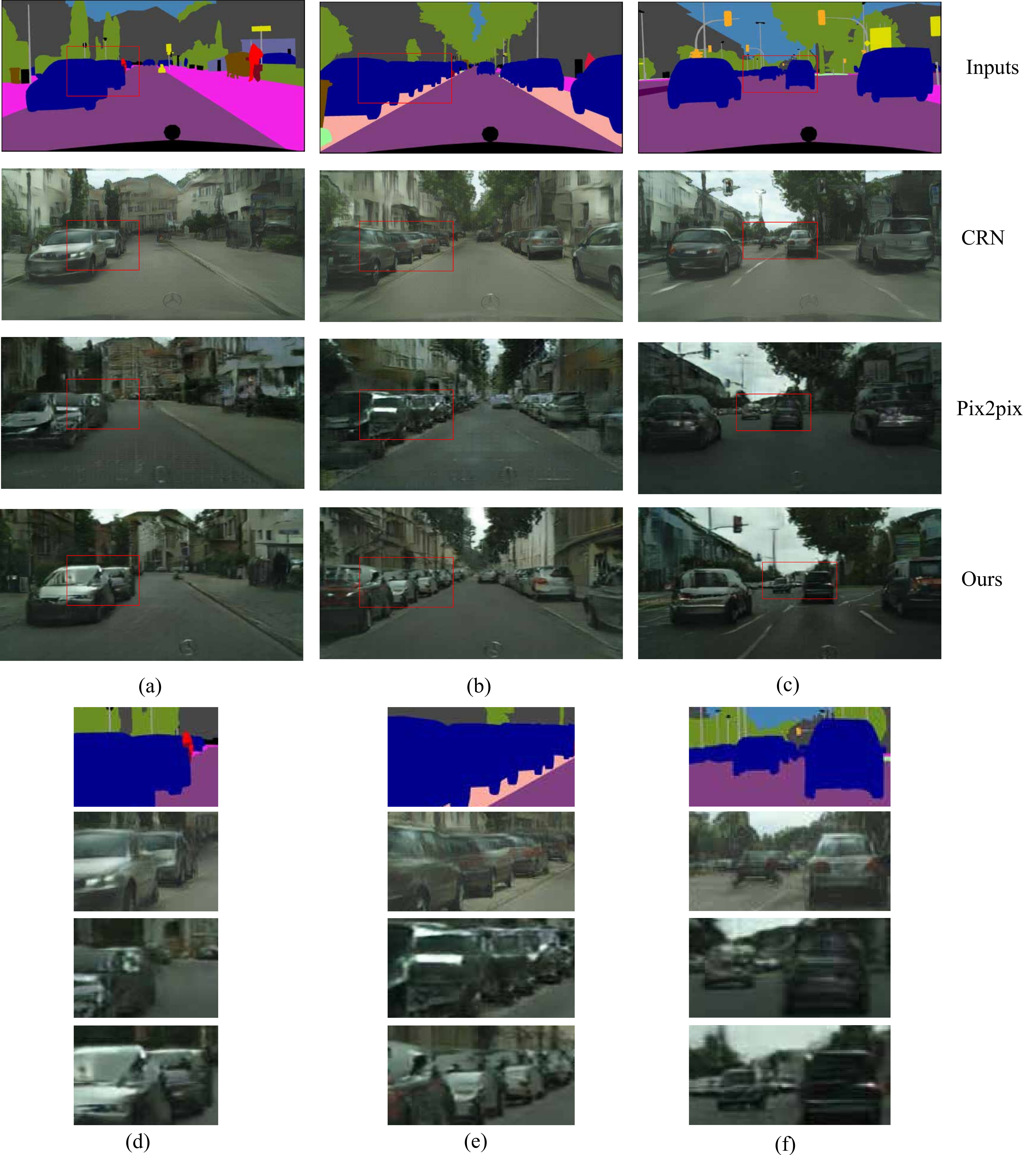}
\caption{Some examples of our methods comparing with Pix2pix \cite{isola2016image} and CRN\cite{chen2017photographic} on Cityscapes dataset. The red boxes show some generation results of overlapped cars.}
\label{fig:city_compare}
\end{figure*}

\begin{table}
\begin{center}
\caption{The image equality evaluation results of different methods on the Cityscapes dataset.}
\label{table:city}
\begin{tabular}{|c|c|c|c|c|}
\hline
Method & P-SNR &MSE &R-MSE &SSIM\\
\hline\hline
Ours & 18.3344 & 0.01677& 0.1250 & 26.0268\\
CRN & 11.9927 & 0.06791 &0.2551 & 16.9266\\
Pix2pix & 15.7161 & 0.03025 & 0.1686 & 16.9876\\
\hline
\end{tabular}
\end{center}

\end{table}

\begin{figure}[t]
\centering
\includegraphics[width=\linewidth]{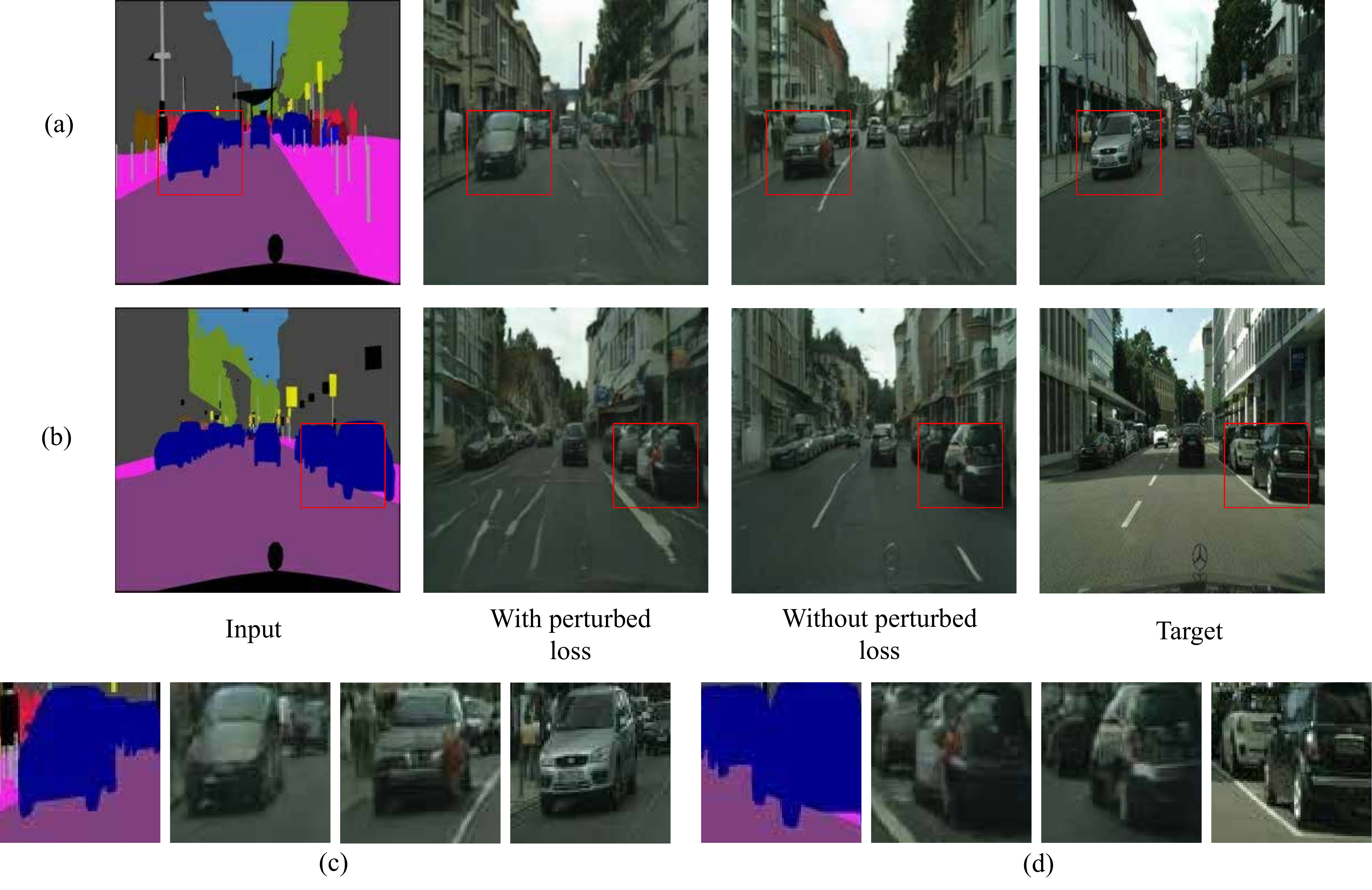}
\caption{More detail of cars can be generated after using pertubed loss .}
\label{fig:lp-city}
\end{figure}

The cityscapes dataset contain 20 different kinds of objects and have various scenes. And the overlapped objects such as cars in the semantic layouts are difficult for our generator to synthesize. As described in Fig. \ref{fig:city_compare}, although models like pix2pix can synthesize a clear image, such generated images often contain mistakes especially near by the boundary of two different objects (e.g. two neighboring cars). The results generated from CRN are basically correct but lack of the details on small objects and the generated objects are blurry. With the help of cascade layer loss, perturbed loss and the instance map, images generated by our model contain correct and clear objects. We also evaluate the image equality scores on the cityscape dataset. The results in Table. \ref{table:city} show that our method performs significantly better than two other methods. And this dataset contains 2975 images and use 2675 for training and others for testing.

\begin{table}
\begin{center}
\caption{The image equality evaluation results of different methods on the Cityscapes dataset.}
\label{table:comparison}
\begin{tabular}{|c|c|c|c|c|}
\hline
Method & P-SNR &MSE &R-MSE &SSIM\\
\hline\hline
Z0.1 & 16.1530 & 0.02823& 0.1617 & 18.1796\\
Z0.04 & 15.7261 & 0.03098 &0.1695 & 16.7215\\
Z0.4 & 15.6197 & 0.03101 & 0.1706 & 16.8597\\
$\mathcal{L}_{p}$ & 16.6002 & 0.02443 & 0.1518 & 21.5974\\
$\mathcal{L}_{c}$ & 16.4890 & 0.02642 & 0.1558 & 20.0253\\
$\mathcal{L}_{c}+ \mathcal{L}_{p}$ & 16.7169 & 0.02386 & 0.1499 & 21.4486\\
$\mathcal{L}_{c}$+ Ins. & 17.7136 & 0.02244 & 0.1451 & 21.3573\\
$\mathcal{L}_{p}$+ Ins. & 16.3767 & 0.02349 & 0.1496 & 22.2922\\
$\mathcal{L}_{c}+ \mathcal{L}_{p}$+ Ins. & 17.4317 & 0.01902 & 0.1236 & 25.4498\\
Ours & 18.3344 & 0.01677& 0.1250 & 26.0268\\
\hline
\end{tabular}
\end{center}

\end{table}

The perturbed loss $\mathcal{L}_{p}$ can improve the image generation in some way. The Fig. ~\ref{fig:lp-city} shows the results we explore the effectiveness of the $\mathcal{L}_{p}$.We see that using the perturbed loss can generate images with more details and the generation got sharper than cases without using perturbed loss. The Table.~\ref{table:comparison} also shows the evaluation results with the four evaluation metric. The perturbed loss allow the adversarial network to train longer efficiently. That's why model with perturbed loss lead to lower MSE error and get better generations. The $\mathcal{L}_{p}$ can help improve the ability of discriminator to distinguish the fake component from the real sample distribution. So it can generate more plausible results.

\begin{figure}[t]
\centering
\includegraphics[width=0.5\textwidth]{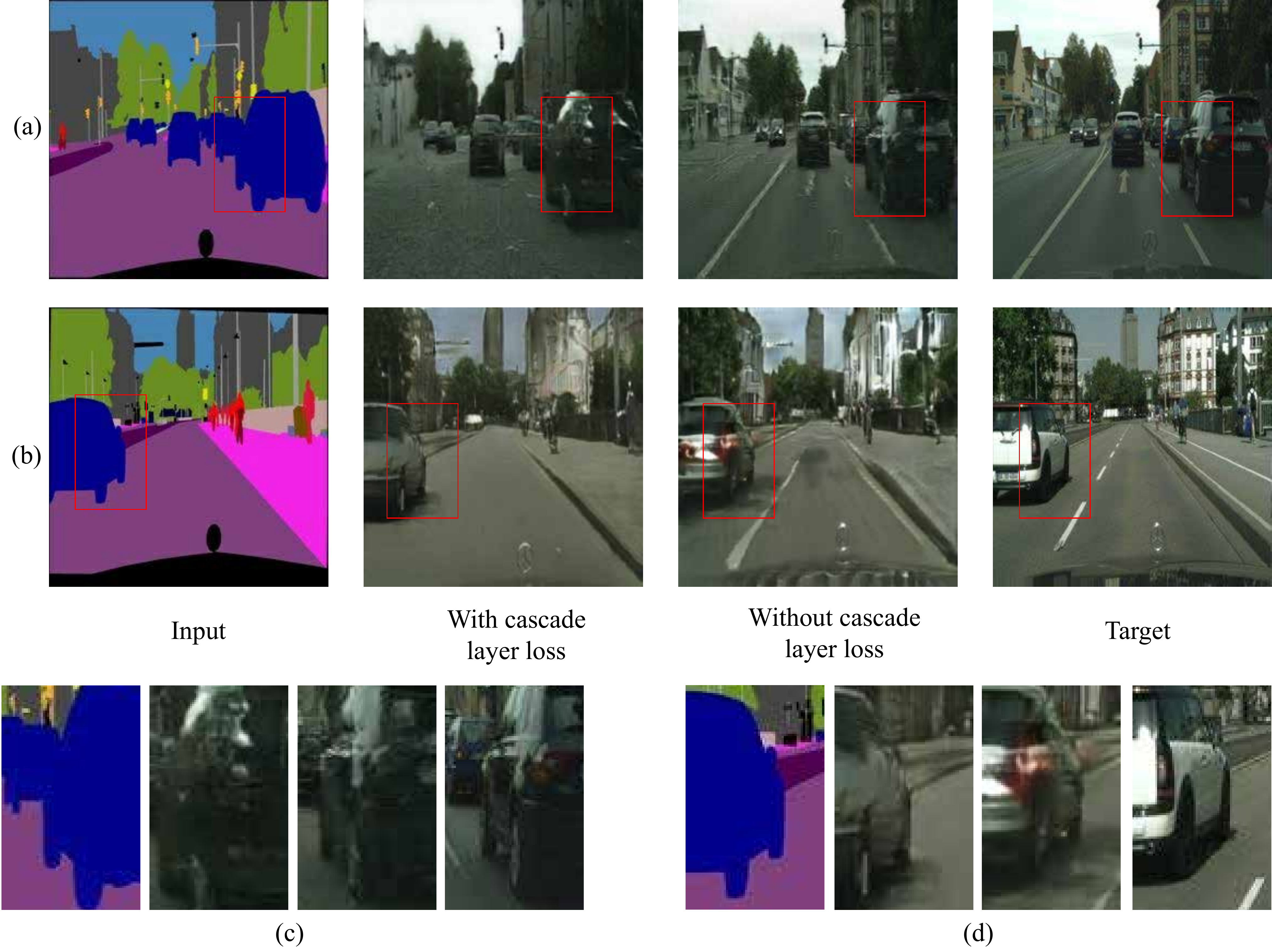}
\caption{The car contents can be well modeled after using cascade loss.}
\label{fig:layer-city}
\end{figure}

And we also consider to check the extra cascade layer loss $\mathcal{L}_{c}$. So we design the contrast experiments to show the effectiveness of this loss. The Fig. ~\ref{fig:layer-city} shows the experimental results. As this figure shows, the results using the cascade layer loss could be clearer. This loss could provide hierarchy constrains for the generator. The higher-layer feature maps can carry on interpretation information for this image-to-image translation. From the Table.~\ref{table:comparison}, the $\mathcal{L}_{c}$ can help model to generate images with higher SSIM and lower MSE error. This loss can force the generator to generate images with considering small parts as well as matching the representations of fake images and real images in high-level space.

\begin{figure}[t]
\centering
\includegraphics[width=0.50\textwidth]{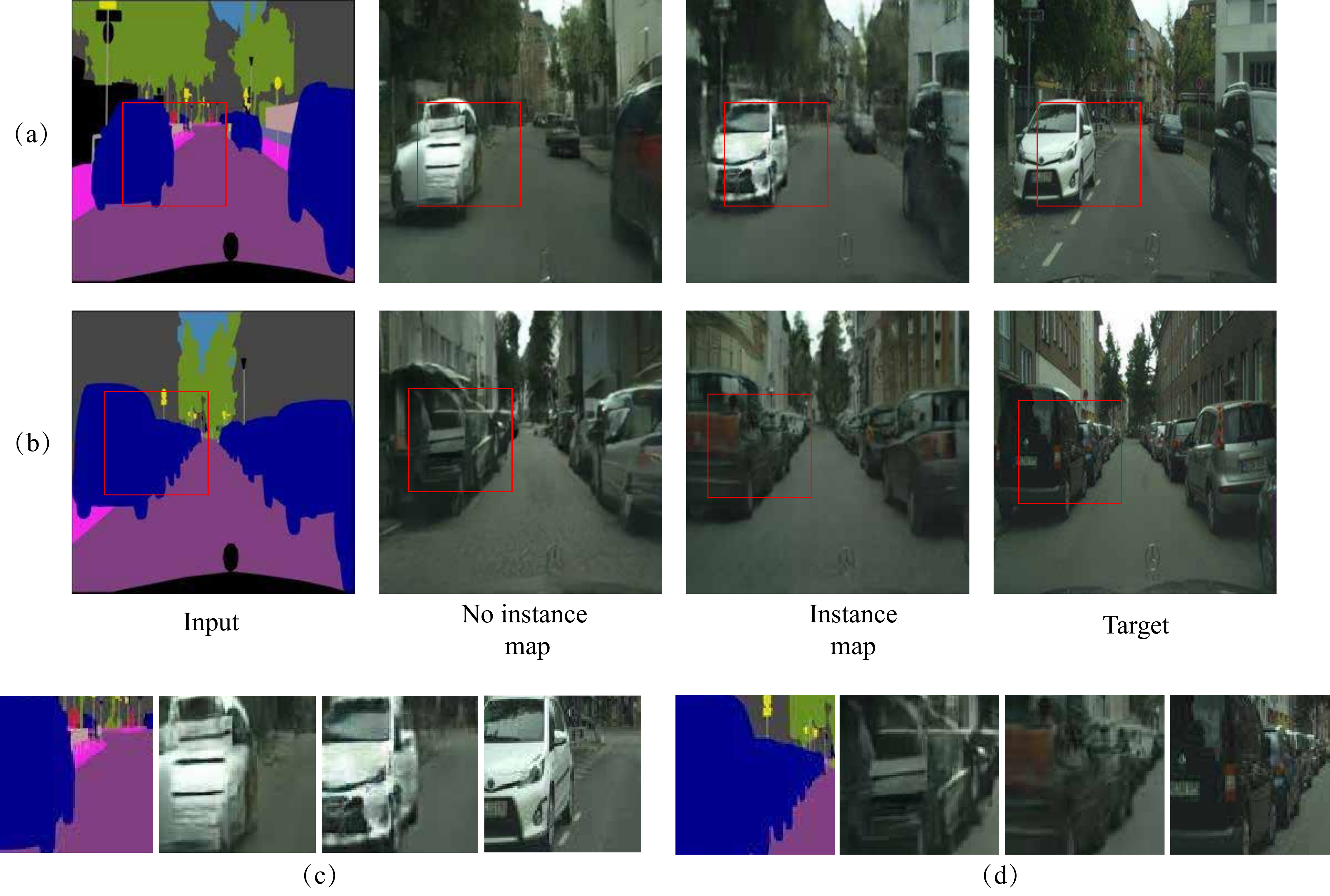}
\caption{The instance map can reduce the risk of wrong contents generation.}
\label{fig:waterconcate}
\end{figure}

The effect of instance map Ins. is also evaluated in our experiment in Table. ~\ref{table:comparison}. To further validate our car instance map Ins. can improve the generation results, we show some examples in the Fig. ~\ref{fig:waterconcate}. The additional car instance map can provide our generator more information and force the discriminator to concentrate on the car objects. It's easy to find that inputs with car instance map lead to more plausible results. Results from Table.~\ref{table:comparison} also demonstrate this point.

\begin{figure}[t]
\centering
\includegraphics[width=0.5\textwidth]{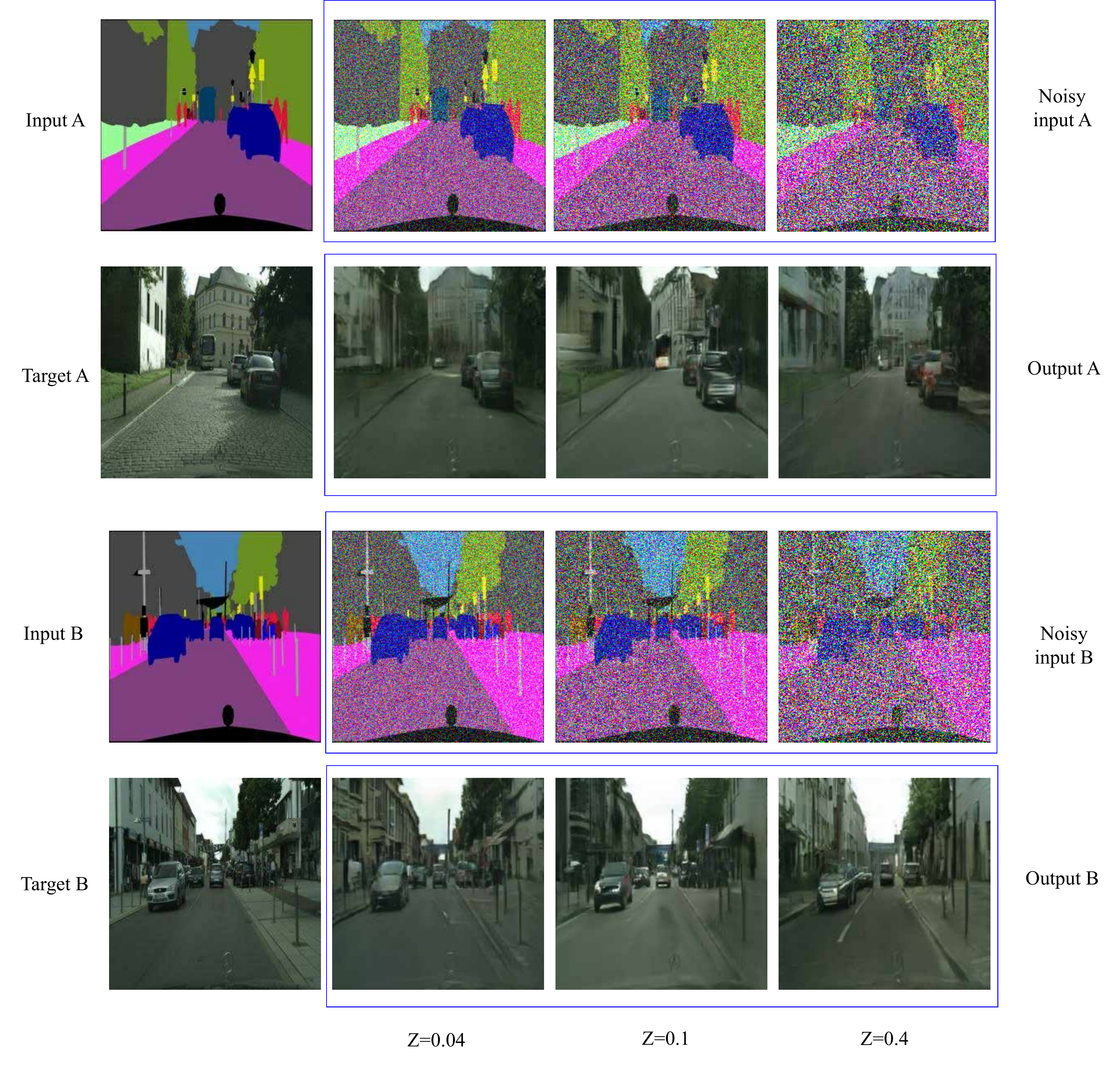}
\caption{Different scale of Gaussian noise induce different generation results on the Cityscape dataset.}
\label{fig:noise}
\end{figure}

In order to explore the effectiveness of noise input, we evaluate how the denoising process can improve the generation results. The Fig. ~\ref{fig:noise} shows some experimenting results. We find that models trained with appropriate Gaussian noise (e.g. Gaussian noise with deviation $0.1$) generate more plausible images. But if the noise is larger (e.g. Gaussian noise with deviation $0.4$), the outputs get worse. We can conclude that an appropriate amount of Gaussian noise can improve the image generation process. But when the noise are too large, the results got worse than results without adding noise. The Table. ~\ref{table:comparison} shows some image equality scores under different condition on the Cityscapes dataset. With a proper noisy input, the denoising process can improve the robustness of our model as well as generating better images.

Considering that we used multiple improvements, we evaluate the some combinations of theses methods on Cityscapes dataset. The Table.~\ref{table:comparison} shows the evaluation results. We find that our framework performs well when we mix Gaussian random noise with standard deviation 0.1 rather than 0.04 and 0.4. The image generation results can be further improved when we consider the perturbed loss $\mathcal{L}_{p}$, the cascade loss $\mathcal{L}_{c}$ and the instance map. We find that we can get better results by hybrid these methods all. We evaluate our final method on the Cityscapes dataset with $\mathcal{L}_{c}$, $\mathcal{L}_{p}$, car instance map and the Gaussian noise with a standard deviation 0.1 and achieved the highest P-SNR and SSIM scores as well as the lowest MSE and R-MSE scores. 


%
%
\section{Conclusion} 
In this paper, we propose a new architecture based on GAN to achieve image translation between two different domains. We introduce a denoising image-to-image framework to improve the image generation process and increase the robustness of our model. We use the instance map to force the generator to concentrate on the overlapped objects to achieve better generation performance. The perturbed loss and the cascade loss can improve the contents organizing ability of the generator. Experimental results show that our method outperforms other state-of-the-art image-to-image translation methods.

Although our model can achieve plausible image-to-image translation tasks among contemporaneous methods, the gap still exists between the real images and the synthesized ones. Generally, the key of image-to-image translation is how to teach the model comprehend the relationship among various objects. On this point, we use cascade layer loss to control the image structure information in different scales. We also use a denoising process and the perturbed loss to improve the robustness and model the invariability in the sample distribution. But they are still not enough to obtain a realistic synthesized image. It's still a challenge to achieve a pixel-level translation. We need the discriminator to be smarter to guide the generation process. It requires the model achieve not only the feature matching, but also the global information extracting. We leave them to our future work.

\section*{Acknowledgment}
This work was supported by the National Natural Science Foundation of China under Grant no. 61701463,  the China Postdoctoral Science Foundation under Grant no. 2017M622277, Natural Science Foundation of Shandong Province of China under Grant no. ZR2017BF011,  the Fundamental Research Funds for the Central Universities under Grant nos. 201713019 and the Qingdao Postdoctoral Science Foundation of China. We gratefully acknowledge the support of NVIDIA Corporation with the donation of the Titan X Pascal GPU used for this research.

\bibliographystyle{IEEEtran}
\bibliography{acess}
\end{document}